\documentstyle[latex-acl]{article}
\title{\vspace{-0.5in}\LARGE\bf RELATING COMPLEXITY TO PRACTICAL
PERFORMANCE IN PARSING WITH WIDE-COVERAGE UNIFICATION GRAMMARS}
\author{John Carroll \\ University of Cambridge, Computer Laboratory \\
Pembroke Street, Cambridge CB2 3QG, UK \\{\it jac@cl.cam.ac.uk}}

\newcommand{\pict}[3]{
 ~ \hfill
 \vbox to #3{
  \hrule width #2 height 0pt depth 0pt
  \vfill
  \special{picture #1}}
 \hfill ~}
\begin{document}

\maketitle
\vspace{-0.5in}
\begin{abstract}

The paper demonstrates that exponential complexities with respect to
grammar size and input length have little impact on the performance of three
unification-based parsing algorithms, using a wide-coverage grammar. The
results imply that the study and optimisation of unification-based parsing
must rely on empirical data until complexity theory can more accurately
predict the practical behaviour of such parsers\footnote{This research was
supported by SERC/DTI project 4/1/1261 `Extensions to the Alvey Natural
Language Tools' and by EC ESPRIT BRA-7315 `ACQUILEX-II'. I am grateful to Ted
Briscoe for comments on an earlier version of this paper,
to David Weir for valuable discussions, and to Hiyan Alshawi for assistance
with the CLARE system.}.
\end{abstract}

\section{1. INTRODUCTION}

General-purpose natural language (NL) analysis systems have recently started
to use declarative unification-based sentence grammar formalisms; systems of
this type include SRI's CLARE system (Alshawi {\it et~al.}, 1992) and the
Alvey NL Tools (ANLT; Briscoe {\it et~al.}, 1987a). Using a declarative
formalism helps ease the task of developing and maintaining the grammar
(Kaplan, 1987). In addition to syntactic processing, the systems
incorporate lexical, morphological, and semantic processing, and have been
applied successfully to the analysis of naturally-occurring texts (e.g.\
Alshawi {\it et~al.}, 1992; Briscoe \& Carroll, 1993).

Evaluations of the grammars in these particular systems have shown them to
have wide coverage (Alshawi {\it et~al.}, 1992; Taylor, Grover \& Briscoe,
1989)\footnote{For example, Taylor {\it et~al.}\ demonstrate that the ANLT
grammar is in principle able to analyse 96.8\% of a corpus of 10,000 noun
phrases taken from a variety of corpora.}. However, although the practical
throughput of parsers with such realistic grammars is important, for example
when processing large amounts of text or in interactive applications, there
is little published research that compares the performance of different
parsing algorithms using wide-coverage unification-based grammars. Previous
comparisons have either focussed on context-free (CF) or augmented CF
parsing (Tomita, 1987; Billot \& Lang, 1989), or have used relatively small,
limited-coverage unification grammars and lexicons (Shann, 1989; Bouma \&
van Noord, 1993; Maxwell \& Kaplan, 1993). It is not clear that these
results scale up to reflect accurately the behaviour of parsers using
realistic, complex unification-based grammars: in particular, with grammars
admitting less ambiguity parse time will tend to increase more slowly with
increasing input length, and also with smaller grammars rule
application can be constrained tightly with relatively simple predictive
techniques. Also, since none of these studies relate observed performance to
that of other comparable parsing systems, implementational oversights may not
be apparent and so be a confounding factor in any general conclusions made.

Other research directed towards improving the throughput of
unification-based parsing systems has been concerned with the unification
operation itself, which can consume up to 90\% of parse time (e.g.\
Tomabechi, 1991) in systems using lexicalist grammar formalisms (e.g.\ HPSG;
Pollard \& Sag, 1987). However, parsing algorithms assume more importance
for grammars having more substantial phrase structure components, such as
CLARE (which although employing some HPSG-like analyses still contains
several tens of rules) and the ANLT (which uses a formalism derived from
GPSG; Gazdar {\it et~al.}, 1985), since the more specific rule set can be
used to control which unifications are performed.

In NL analysis, the syntactic information associated with lexical items makes
top-down parsing less attractive than bottom-up (e.g.\ CKY; Kasami, 1965;
Younger, 1967), although the latter is often augmented with top-down
prediction to improve performance (e.g.\ Earley, 1970; Lang, 1974; Pratt,
1975). Section 2 describes three unification-based parsers which are related
to polynomial-complexity bottom-up CF parsing algorithms. Although
incorporating unification increases their complexity to exponential on grammar
size and input length (section 3), this appears to have little impact on
practical performance (section 4). Sections 5 and 6 discuss these findings
and present conclusions.

\section{2. THE PARSERS}

The three parsers in this study are: a bottom-up left-corner parser, a
(non-deterministic) LR parser, and an LR-like parser based on an algorithm
devised by Schabes (1991). All three parsers accept grammars written in the
ANLT formalism (Briscoe {\it et~al.}, 1987a), and the first two are
distributed as part of the ANLT package. The parsers create parse forests
(Tomita, 1987) that incorporate subtree sharing (in which identical
sub-analyses are shared between differing superordinate analyses) and node
packing (where sub-analyses covering the same portion of input whose root
categories are in a subsumption relationship are merged into a single node).

\subsection{THE BOTTOM-UP LEFT-CORNER PARSER}

The bottom-up left-corner (BU-LC) parser operates left-to-right and
breadth-first, storing partial (active) constituents in a chart; Carroll
(1993) gives a full description. Although pure bottom-up parsing is
not usually thought of as providing high performance, the actual implementation
achieves very good throughput (see section 4) due to a number of significant
optimisations, amongst which are:
\begin{itemize}
\item Efficient rule invocation from cheap (static) rule indexing, using
discrimination trees keyed on the feature values in each rule's first
daughter to interleave rule access with unification and also to share
unification results across groups of rules.
\item Dynamic indexing of partial and complete constituents on category types
to avoid attempting unification or subsumption operations which static analysis
shows will always fail.
\item Dynamic storage minimisation, deferring structure copying---e.g.\
required by the unification operation or by constituent creation---until
absolutely necessary (e.g.\ unification success or parse success,
respectively).
\end{itemize}
The optimisations improve throughput by a factor of more than three.

\subsection{THE NON-DETERMINISTIC LR PARSER}

Briscoe \& Carroll (1993) describe a methodology for constructing an LR parser
for a unification-based grammar, in which a CF `backbone' grammar is
automatically constructed from the unification grammar, a parse table is
constructed from the backbone grammar, and a parser is driven by the table and
further controlled by unification of the `residue' of features in the
unification grammar that are not encoded in the backbone. In this parser, the
LALR(1) technique (Aho, Sethi \& Ullman, 1986) is used, in conjunction with a
graph-structured stack (Tomita, 1987), adapting for unification-based parsing
Kipps' (1989) Tomita-like recogniser that achieves polynomial complexity on
input length through caching.

On each reduction the parser performs the
unifications specified by the unification grammar version of the CF backbone
rule being applied. This constitutes an on-line parsing algorithm. In the
general case, the off-line variant (in which all unifications are deferred
until the complete CF parse forest has been constructed) is not guaranteed to
terminate; indeed, it usually does not do so with the ANLT grammar. However, a
drawback to the on-line algorithm is that a variant of Kipps' caching cannot
be used, since the cache must necessarily assume that all reductions at a given
vertex with all rules with the same number of daughters build exactly
the same constituent every time; in general this is not the case when the
daughters are unification categories. A weaker kind of cache on
partial analyses (and thus unification results) was found to be necessary in
the implementation, though, to avoid duplication of unifications; this sped
the parser up by a factor of about three, at little space cost.

\subsection{THE COMPILED-EARLEY PARSER}

The Compiled-Earley (CE) parser is based on a predictive chart-based CF
parsing algorithm devised by Schabes (1991) which is driven by a table
compiling out the predictive component of Earley's (1970) parser. The size of
the table is related linearly to the size of the grammar (unlike the LR
technique). Schabes demonstrates that this parser always takes fewer steps
than Earley's, although its time complexity is the same: $O(n^{3})$. The space
complexity is also cubic, since the parser uses Earley's representation of
parse forests.

The incorporation of unification into the CE parser follows the methodology
developed for unification-based LR parsing described in the previous section:
a table is computed from a CF `backbone', and a parser, augmented with on-line
unification and feature-based subsumption operations, is driven by the table.
To allow meaningful comparison with the LR parser, the CE parser uses a
one-word lookahead version of the table, constructed using a modified LALR
technique (Carroll, 1993)\footnote{Schabes describes a table with no
lookahead; the successful application of this technique supports Schabes'
(1991:109) assertion that ``several other methods (such as LR(k)-like and
SLR(k)-like) can also be used for constructing the parsing tables [...]''}.

To achieve the cubic time bound, the parser must be able to retrieve in unit
time all items in the chart having a given state, and start and end position in
the input string. However, the obvious array implementation, for say a ten
word sentence with the ANLT grammar, would contain almost 500000 elements. For
this reason, the implementation employs a sparse representation for the array,
since only a small proportion of the elements are ever
filled. In this parser, the same sort of duplication of unifications occurs as
in the LR parser, so lists of partial analyses are cached in the same way.

\section{3. COMPLEXITIES OF THE PARSERS}

The two variables that determine a parser's computational complexity are the
grammar and the input string (Barton, Berwick \& Ristad, 1987).
These are considered separately in the next two sections.

\subsection{GRAMMAR-DEPENDENT COMPLEXITY}

The term dependent on the grammar in the time complexity of the BU-LC
unification-based parser described above is $O(|C|^{2}|R|^{3})$, where $|C|$
is the number of categories implicit in the grammar, and $|R|$, the number of
rules. The space complexity is
dominated by the size of the parse forest, $O(|C|)$ (these results are proved
by Carroll, 1993). For the ANLT grammar, in which features are nested to a
maximum depth of two, $|C|$ is finite but nevertheless extremely large
(Briscoe {\it et~al.}, 1987b)\footnote{Barton, Berwick \& Ristad (1987:221)
calculate that GPSG, also with a maximum nesting depth of two, licences more
than $10^{775}$ distinct syntactic categories. The number of categories is
actually infinite in grammars that use a fully recursive feature system.}.

The grammar-dependent complexity of the LR parser makes it also appear
intractable: Johnson (1989) shows that the number of LR(0) states for certain
(pathological) grammars is exponentially related to the size of the grammar,
and that there are some inputs which force an LR parser to visit all of these
states in the course of a parse. Thus the total number of operations
performed, and also space consumed (by the vertices in the graph-structured
stack), is an exponential function of the size of the grammar.

To avoid this complexity, the CE parser employs a table construction method
which ensures that the number of states in the parse table is linearly related
to the size of the grammar, resulting in the number of operations performed by
the parser being at worst a polynomial function of grammar size.

\subsection{INPUT-DEPENDENT COMPLEXITY}

Although the complexity of returning all parses for a string is always
related exponentially to its length (since the number of parses
is exponential, and they must all at least be enumerated), the complexity of a
parser is usually measured for the computation of a parse forest (unless
extracting a single analysis from the forest is worse than
linear)\footnote{This complexity measure does correspond to real world
usage of a parser, since practical systems can usually afford to extract
only a small number of parses from the frequently very large number encoded in
a forest; this is often done on the basis of preference-based or probabilistic
factors (e.g.\ Carroll \& Briscoe, 1992).}.

If one of the features of the ANLT grammar formalism, the kleene operator
(allowing indefinite repetition of rule daughters), is disallowed, then the
complexity of the BU-LC parser with respect to the length of the input string
is $O(n^{\rho+1})$, where $\rho$ is the maximum number of daughters in a rule
(Carroll, 1993). The inclusion of the operator increases the complexity to
exponential. To retain the polynomial time bound, new rules can be introduced
to produce recursive tree structures instead of an iterated flat tree
structure. However, when this technique is applied to the ANLT grammar the
increased overheads in rule invocation and structure building actually slow
the parser down.

Although the time and space complexities of CF versions of the LR and CE
parsers are $O(n^{3})$, the unification versions of these parsers both turn
out to have time bounds that are greater than cubic, in the general case. The
CF versions implicitly pack identical {\it sequences} of sub-analyses, and in
all reductions at a given point with rules with the same number of daughters,
the packed sequences can be formed into higher-level constituents as they
stand without further processing. However,
in the unification versions, on each reduce action the daughters of the rule
involved have to be unified with every possible alternative sequence of the
sub-analyses that are being consumed by the rule (in effect expanding and
flattening out the packed sequences), leading to a bound of
$n^{\rho+1}$ on the total number of unifications.

\section{4. PRACTICAL RESULTS}

To assess the practical performance of the three unification-based parsers
described above, a series of experiments were conducted using the ANLT grammar
(Grover, Carroll \& Briscoe, 1993), a wide-coverage grammar of English. The
grammar is defined in metagrammatical formalism which is compiled into a
unification-based `object grammar'---a syntactic variant of the Definite
Clause Grammar formalism (Pereira \& Warren, 1980)---containing 84 features
and 782 phrase structure rules. Parsing uses fixed-arity term unification. The
grammar provides full coverage of the following constructions: declarative
sentences, imperatives and questions (yes/no, tag and wh-questions); all
unbounded dependency types (topicalisation, relativisation, wh-questions); a
relatively exhaustive treatment of verb and adjective complement types;
phrasal and prepositional verbs of many complement types; passivisation; verb
phrase extraposition; sentence and verb phrase modification; noun phrase
complements and pre- and post-modification; partitives; coordination of all
major category types; and nominal and adjectival comparatives.

Although the grammar is linked to a lexicon containing definitions for 40000
base forms of words, the experiments draw on a much smaller lexicon of 600
words (consisting of closed class vocabulary and, for open-class vocabulary,
definitions of just a sample of words which taken together exhibit the full
range of possible complementation patterns), since issues of lexical coverage
are of no concern here.

\subsection{COMPARING THE PARSERS}

In the first experiment, the ANLT grammar was
loaded and a set of sentences was input to each of the three parsers. In
order to provide an independent basis for comparison, the same sentences were
also input to the SRI Core Language Engine (CLE) parser (Moore \& Alshawi,
1992) with the CLARE2.5 grammar (Alshawi {\it et~al.}, 1992), a
state-of-the-art system accessible to the author.

The sentences were taken from an initial sample of 175 representative
sentences extracted from a corpus of approximately 1500 that form
part of the ANLT package. This corpus, implicitly defining the types of
construction the grammar is intended to cover, was written by the linguist
who developed the ANLT grammar and is used to check for any adverse effects on
coverage when the grammar is modified during grammar development. Of the
initial 175 sentences, the CLARE2.5 grammar failed to parse 42 (in several
cases because punctuation is strictly required but is missing from the
corpus). The ANLT grammar also failed to parse three of these, plus an
additional four. These sentences were removed from the sample, leaving 129
(mean length 6.7 words) of which 47 were
declarative sentences, 38 wh-questions and other sentences with gaps, 20
passives, and 24 sentences containing co-ordination.

Table~1 shows the total parse times and storage allocated for the BU-LC
parser, the LR parser, and the CE parser, all with ANLT grammar
and lexicon.
\begin{table}
\centering
\begin{tabular}{|l|l|r|r|} \hline
Parser	& Grammar	 & CPU time	& Storage\\
       &          &          & allocated\\ \hline
BU-LC	 & ANLT	    & 75.5	    & 47.0\\
LR	    & ANLT	    & 48.9	    & 33.6\\
CE     & ANLT     & 98.4     & 38.5\\
& & & \\
CLE	   & CLARE2.5	& 277.7	   & --\\ \hline
\end{tabular}
\caption{Parse times (in CPU seconds on a Sun Sparc ELC workstation) and
storage allocated (in megabytes) while parsing the 129 test sentences
(1--12 words in length).}
\end{table}
All three parsers have been implemented by the author to a similar
high standard: similar implementation techniques are used in all the parsers,
the parsers share the same unification module, run in the same Lisp
environment, have been compiled with the same optimisation settings, and
have all been profiled with the same tools and hand-optimised to a similar
extent. (Thus any difference in performance of more than around 15\% is
likely to stem from algorithmic rather than implementational reasons). Both of
the predictive parsers employ one symbol of lookahead, incorporated into the
parsing tables by the LALR technique. Table 1 also shows the results for the
CLE parser with the CLARE2.5 grammar and lexicon. The figures include garbage
collection time, and phrasal (where appropriate) processing, but not parse
forest unpacking. Both grammars give a total of around 280 analyses at a
similar level of detail.

The results show that the LR parser is approximately 35\% faster than the BU-LC
parser, and allocates about 30\% less storage. The magnitude of the speed-up is
less than might be expected, given the enthusiastic advocation of
non-deterministic CF LR parsing for NL by some researchers (e.g.\ Tomita,
1987; Wright, Wrigley \& Sharman, 1991), and in the light of improvements
observed for predictive over pure bottom-up parsing (e.g.\ Moore \& Dowding,
1991). However, on the assumption that incorrect prediction of gaps is the main
avoidable source of performance degradation (c.f.\ Moore \& Dowding), further
investigation shows that the speed-up is near the maximum that is possible
with the ANLT grammar (around 50\%).

The throughput of the CE parser is half that of the LR parser, and
also less than that of the BU-LC parser. However, it is intermediate between
the two in terms of storage allocated. Part of the difference in performance
between it and the LR parser is due to the fact that it performs around 15\%
more unifications. This might be expected since the corresponding finite state
automaton is not determinised---to avoid theoretical exponential time
complexity on grammar size---thus paying a price at run time. Additional
reasons for the relatively poor performance of the CE parser are the
overheads involved in maintaining a sparse representation of the chart, and
the fact that with the ANLT grammar it generates less ``densely packed'' parse
forests, since its parse table, with 14\% more states (though
fewer actions) than the LALR(1) table, encodes more contextual distinctions
(Billot \& Lang, 1989:146).

Given that the ANLT and CLARE2.5 grammars have broadly similar (wide) coverage
and return very similar numbers of syntactic analyses for the same inputs, the
significantly better throughput of the three parsers described in this paper
over the CLE parser\footnote{Although the ANLT parser is implemented in Common
Lisp and the CLE parser in Prolog, comparing parse times is a valid exercise
since current compiler and run-time support technologies for both languages are
quite well-developed, and in fact the CLE parser takes advantage of Prolog's
built-in unification operation which will have been very tightly coded.}
indicates that they do not contain any significant implementational
deficiencies which would bias the results\footnote{The ANLT's speed advantage
over CLARE is less pronounced if the time for morphological analysis and
creation of logical forms is taken into account, probably because the systems
use different processing techniques in these modules.}.

\subsection{SWAPPING THE GRAMMARS OVER}

A second experiment was carried out with the CLE parser, in which the
built-in grammar and lexicon were replaced by versions of the ANLT object
grammar and lexical entries translated (automatically) into the CLE formalism.
(The reverse of this configuration, in which the CLARE2.5 grammar is translated
into the ANLT formalism, is not possible since some central rules contain
sequences of daughters specified by a single `list' variable, which has no
counterpart in the ANLT and cannot directly be simulated). The throughput of
this configuration was only one fiftieth of that of the BU-LC parser. The ANLT
grammar contains more than five times as many rules than does the
sentence-level portion of the CLARE2.5 grammar, and Alshawi (personal
communication) points out that the CLE parser had not previously been run
with a grammar containing such a large number of rules, in contrast to the ANLT
parsers.

\subsection{THE EFFECT OF SENTENCE LENGTH}

Although the mean sentence length in the first two experiments is much shorter
than the 20--30 word length (depending on genre etc.)\ that is common in real
texts, the test sentences cover a wide range of syntactic constructions and
exhibit less constructional bias than would a set of sentences extracted at
random from a single corpus. However, to investigate performance on longer
sentences and the relationship between sentence length and parse time, a
further set of 100 sentences with lengths distributed uniformly between 13 and
30 words was created by hand by the author and added to the previous test
data. Table~2 shows the relationship between sentence length and mean parse
time with the BU-LC and LR parsers.
\begin{table*}[t]
\centering
\begin{tabular}{|r|r|r|r|r|r|r|} \hline
\multicolumn{1}{|c|}{Sentence}	&
\multicolumn{2}{c|}{BU-LC} &
\multicolumn{2}{c|}{LR} &
\multicolumn{2}{c|}{Number of}\\
\multicolumn{1}{|c|}{length}	&
\multicolumn{2}{c|}{Parse time} &
\multicolumn{2}{c|}{Parse time} &
\multicolumn{2}{c|}{parses}\\
\multicolumn{1}{|c|}{(words)}&
\multicolumn{1}{c|}{Mean} &
\multicolumn{1}{c|}{$\sigma$} &
\multicolumn{1}{c|}{Mean} &
\multicolumn{1}{c|}{$\sigma$} &
\multicolumn{1}{c|}{Mean} &
\multicolumn{1}{c|}{$\sigma$}\\ \hline
1--3 & 0.11 & 0.06 & 0.05 & 0.02 & 1.3 & 0.7\\
4--6 & 0.23 & 0.18 & 0.15 & 0.11 & 1.4 & 0.8\\
7--9 & 0.42 & 0.24 & 0.28 & 0.17 & 1.8 & 1.3\\
10--12 & 1.17 & 0.92 & 0.76 & 0.52 & 3.8 & 2.4\\
13--15 & 0.97 & 0.28 & 0.86 & 0.38 & 10.0 & 13.7\\
16--18 & 1.92 & 0.75 & 1.89 & 1.00 & 14.3 & 17.5\\
19--21 & 3.54 & 1.42 & 3.74 & 2.46 & 60.1 & 117.3\\
22--24 & 3.87 & 1.62 & 3.61 & 3.07 & 143.8 & 200.1\\
25--27 & 5.45 & 1.98 & 5.05 & 3.59 & 168.8 & 303.1\\
28--30 & 7.86 & 2.37 & 12.89 & 5.65 & 343.5 & 693.7\\ \hline
\end{tabular}
\caption{Mean and standard deviation parse times (in CPU seconds on an
HP9000/710 workstation), and numbers of parses for the 229 test sentences
(1--30 words in length) with the BU-LC and LR parsers.}
\end{table*}
\begin{figure*}[t]
\centering
\pict{graph}{135mm}{86mm}
\caption{Mean parse times (in CPU seconds on an HP9000/710 workstation) for
the test sentences with the BU-LC and LR parsers. A quadratic function is also
displayed.}
\end{figure*}

In contrast to the results from the first experiment, the throughput of the LR
parser is only 4\% better than that of the BU-LC parser for sentences of 13--27
words in length. The former parses many sentences up to twice as fast, but a
small proportion of the others are parsed almost twice as slowly.
As well as their wide variability with respect to the BU-LC parser, the
absolute variability of the LR parse times is high (reflected in large
standard deviations---$\sigma$---see Table 2). Most of the sentences
for which LR performance is worse contain more than one occurrence of
the passive construction: due to their length this is particularly the case for
the group of sentences of 28--30 words with which the LR parser
performed particularly badly. However, it is likely that if the
constraining power of the parse table were improved in this area the
difference in throughput between LR and BU-LC would revert to nearer the 35\%
figure seen in the first experiment.

The standard deviations for numbers of parses are also relatively large. The
maximum number of parses was 2736 for one 29-word sentence, but on the
other hand some of even the longest sentences had fewer than ten parses. (But
note that since the time taken for parse forest unpacking is not
included in parse times, the latter do not vary by such a large magnitude).

The results of this experiment are displayed graphically in Figure~1,
together with a quadratic function. Comparison with the function suggests that,
at least for the BU-LC parser, parse time is related roughly quadratically to
input length.

In previous work with the ANLT (Briscoe \& Carroll, 1993), throughput with raw
corpus data was worse than that observed in these experiments, though probably
only by a constant factor. This could be due to the fact that the vocabulary
of the corpus concerned exhibits significantly higher lexical ambiguity;
however, for sentences taken from a specific corpus, constructional bias
observed in a training phase could be exploited to improve performance (e.g.\
Samuelsson \& Rayner, 1991).

\section{5. DISCUSSION}

All three of the parsers have theoretical worst-case complexities that are
either exponential, or polynomial on grammar size but with an extremely
large multiplier. Despite this, in the practical experiments reported in the
previous section the parsers achieve relatively good throughput with a
general-purpose wide-coverage grammar of a natural language. It therefore
seems likely that grammars of the type considered in this paper (i.e.\ with
relatively detailed phrase structure components, but comparatively simple from
a unification perspective), although realistic, do not bring the parsing
algorithms involved anywhere near the worst-case complexity.

In the experiments, the CE technique results in a parser with worse
performance than the normal LR technique. Indeed, for the ANLT grammar, the
number of states---the term that the CE technique reduces from exponential to
linear on the grammar size---is actually smaller in the
standard LALR(1) table. This suggests that, when considering the complexity of
parsers, the issue of parse table size is of minor importance for realistic
NL grammars (as long as an implementation represents the table compactly), and
that improvements to complexity results with respect to grammar size, although
interesting from a theoretical standpoint, may have little practical relevance
for the processing of natural language.

Although Schabes (1991:107) claims that the problem of exponential grammar
complexity ``is particularly acute for natural language processing since in
this context the input length is typically small (10--20 words) and the
grammar size very large (hundreds or thousands of rules and symbols)'', the
experiments indicate that, with a wide-coverage NL grammar, inputs of this
length can be parsed quite quickly; however,
longer inputs (of more than about 30 words in length)---which occur relatively
frequently in written text---are a problem. Unless grammar size takes on
proportionately much more significance for such longer inputs, which seems
implausible, it appears that in fact the major problems do not lie in the area
of grammar size, but in input length.

All three parsers have worst-case complexities that are exponential on input
length. This theoretical bound might suggest that parsing performance would be
severely degraded on long sentences; however, the relationship between length
of sentence and parse time with the ANLT grammar and the sentences tested
appears to be approximately only quadratic. There are
probably many reasons why performance is much better than the complexity
results suggest, but the most important may be that:
\begin{itemize}
\item kleene star is used only in a very limited context (for the analysis
of coordination),
\item more than 90\% of the rules in the grammar have no more than two
daughters, and
\item very few rules license both left and right recursion (for instance of
the sort that is typically used to analyse noun compounding, i.e.\ {\tt
N~-->~N~N}).
\end{itemize}

Despite little apparent theoretical difference between the CLE and ANLT
grammar formalisms, and the fact that no explicit or formal process of
`tuning' parsers and grammars to perform well with each other has been carried
out in either of the ANLT or CLARE systems, the results of the experiment
comparing the performance of the respective parsers using the ANLT grammar
suggests that the parallel development of the software and grammars that has
occurred nevertheless appears to have caused this to happen automatically. It
therefore seems likely that implementational decisions and optimisations based
on subtle properties of specific grammars can, and may very often be, more
important than worst-case complexity when considering the practical
performance of parsing algorithms.

\section{6. CONCLUSIONS}

The research reported is in a similar vein to that of, for example, Moore \&
Dowding (1991), Samuelsson \& Rayner (1991), and Maxwell \& Kaplan (1993), in
that it relies on empirical results for the study and optimisation of parsing
algorithms rather than on traditional techniques of complexity analysis. The
paper demonstrates that research in this area will have to rely on empirical
data until complexity theory is developed to a point where it is sufficiently
fine-grained and accurate to predict how the properties of individual
unification-based grammars will interact with particular parsing algorithms to
determine practical performance.

\section*{REFERENCES}
\newcommand{\book}[4]{\item #1 (#4) {\it #2}.\ #3.}
\newcommand{\barticle}[7]{\item #1 (#7) ``#2.''\ In {\it #4}, edited by #5,
#3.\ #6.}
\newcommand{\jarticle}[6]{\item #1 (#6) ``#2.''\ {\it #3,} #4:\ #5.}
\newcommand{\particle}[6]{\item #1 (#6) ``#2.''\ In {\it Proceedings of the
#3}.\ #5.}
\newcommand{\lazyjarticle}[4]{\item #1 (#4) ``#2.''\ {\it #3}.}

\begin{list}{}
   {\leftmargin 1.6em
    \itemindent -\leftmargin
    \itemsep 0pt plus 1pt
    \parsep 0pt plus 1pt}

\book{Aho, A., R. Sethi \& J. Ullman}
{Compilers: principles, techniques and tools}
{Reading, MA: Addison-Wesley}
{1986}

\book{Alshawi, H., D. Carter, R. Crouch, S. Pulman, M. Rayner \& A. Smith}
{CLARE: a contextual reasoning and cooperative response framework for
the Core Language Engine}
{SRI International, Cambridge, UK}
{1992}

\book{Barton, G., R. Berwick \& E. Ristad}
{Computational complexity and natural language}
{Cambridge, MA: MIT Press}
{1987}

\particle{Billot, S. \& B. Lang}
{The structure of shared forests in ambiguous parsing}
{27th Meeting of the Association for Computational Linguistics}
{Vancouver, Canada}
{143--151}
{1989}

\particle{Bouma, G. \& G. van Noord}
{Head-driven parsing for lexicalist grammars: experimental results}
{6th Conference of the European Chapter of the Association for Computational
Linguistics}
{}
{101--105}
{1993}

\particle{Briscoe, E., C. Grover, B. Boguraev \& J. Carroll}
{A formalism and environment for the development of a large grammar of
English}
{10th International Joint Conference on Artificial Intelligence}
{Milan, Italy}
{703--708}
{1987a}

\barticle{Briscoe, E., C. Grover, B. Boguraev \& J. Carroll}
{Feature defaults, propagation and reentrancy}
{Centre for Cognitive Science, Edinburgh University, UK}
{Categories, Polymorphism and Unification}
{E. Klein \& J. van Benthem}
{19--34}
{1987b}

\jarticle{Briscoe, E. \& J. Carroll}
{Generalised probabilistic LR parsing of natural language (corpora) with
unification-based grammars}
{Computational Linguistics}
{19(1)}
{25--59}
{1993}

\book{Carroll, J.}
{Practical unification-based parsing of natural language}
{Computer Laboratory, Cambridge University, UK, Technical Report 314}
{1993}

\particle{Carroll, J. \& E. Briscoe}
{Probabilistic normalisation and unpacking of packed parse
forests for unification-based grammars}
{AAAI Fall Symposium on Probabilistic Approaches to Natural Language}
{Cambridge, MA}
{33--38}
{1992}

\jarticle{Earley, J.}
{An efficient context-free parsing algorithm}
{Communications of the ACM}
{13.2}
{94--102}
{1970}

\book{Gazdar, G., E. Klein, G. Pullum \& I. Sag}
{Generalized phrase structure grammar}
{Oxford, UK: Blackwell}
{1985}

\book{Grover, C., J. Carroll \& E. Briscoe}
{The Alvey natural language tools grammar (4th release)}
{Computer Laboratory, Cambridge University, UK, Technical Report 284}
{1993}

\particle{Johnson, M.}
{The computational complexity of Tomita's algorithm}
{1st International Workshop on Parsing Technologies}
{Carnegie-Mellon University, Pittsburgh}
{203--208}
{1989}

\barticle{Kaplan, R.}
{Three seductions of computational psycholinguistics}
{New York: Academic Press}
{Linguistic Theory and Computer Applications}
{P. Whitelock {\it et~al.}}
{149--188}
{1987}

\book{Kasami, J.}
{An efficient recognition and syntax analysis algorithm for context-free
languages}
{Air Force Cambridge Research Laboratory, Bedford, MA, Report AFCRL-65-758}
{1965}

\particle{Kipps, J.}
{Analysis of Tomita's algorithm for general context-free parsing}
{1st International Workshop on Parsing Technologies}
{Carnegie-Mellon University, Pittsburgh}
{193--202}
{1989}

\barticle{Lang, B.}
{Deterministic techniques for efficient non-deterministic parsers}
{Berlin, Germany: Springer-Verlag}
{Automata, Languages and Programming, Lecture Notes in Computer Science 14}
{J. Loeckx}
{255--269}
{1974}

\jarticle{Maxwell, J. III \& R. Kaplan}
{The interface between phrasal and functional constraints}
{Computational Linguistics}
{19(4)}
{571--590}
{1993}

\barticle{Moore, R. \& H. Alshawi}
{Syntactic and semantic processing}
{Cambridge, MA: MIT Press}
{The Core Language Engine}
{H. Alshawi}
{129--148}
{1992}

\particle{Moore, R. \& J. Dowding}
{Efficient bottom-up parsing}
{DARPA Speech and Natural Language Workshop}
{Asilomar, CA}
{200--203}
{1991}

\jarticle{Pereira, F. \& D. Warren}
{Definite clause grammars for language analysis---a survey of the formalism and
a comparison with augmented transition networks}
{Artificial Intelligence}
{13(3)}
{231--278}
{1980}

\book{Pollard, C. \& I. Sag}
{Information-based syntax and semantics: volume 1--fundamentals}
{Chicago, IL: University of Chicago Press}
{1987}

\particle{Pratt, V.}
{LINGOL -- a progress report}
{5th International Joint Conference on Artificial Intelligence}
{Tbilisi, USSR}
{422--428}
{1975}

\particle{Samuelsson, C. \& M. Rayner}
{Quantitative evaluation of explanation-based learning as an optimization tool
for a large-scale natural language system}
{12th International Joint Conference on Artificial Intelligence}
{Sydney, Australia}
{609--615}
{1991}

\particle{Schabes, Y.}
{Polynomial time and space shift-reduce parsing of arbitrary context-free
grammars}
{29th Annual Meeting of the Association for Computational Linguistics}
{Berkeley, CA}
{106--113}
{1991}

\particle{Taylor, L., C. Grover \& E. Briscoe}
{The syntactic regularity of English noun phrases}
{4th European Meeting of the Association for Computational Linguistics}
{UMIST, Manchester, UK}
{256--263}
{1989}

\particle{Tomabechi, H.}
{Quasi-destructive graph unification}
{29th Annual Meeting of the Association for Computational Linguistics}
{Berkeley, CA}
{315--322}
{1991}

\jarticle{Tomita, M.}
{An efficient augmented-context-free parsing algorithm}
{Computational Linguistics}
{13(1)}
{31--46}
{1987}

\particle{Shann, P.}
{The selection of a parsing strategy for an on-line machine translation system
in a sublanguage domain. A new practical comparison}
{1st International Workshop on Parsing Technologies}
{Pittsburgh, PA}
{264--276}
{1989}

\particle{Wright, J., E. Wrigley \& R. Sharman}
{Adaptive probabilistic generalized LR parsing}
{2nd International Workshop on Parsing Technologies}
{Cancun, Mexico}
{154--163}
{1991}

\jarticle{Younger, D.}
{Recognition and parsing of context-free languages in time $n^{3}$}
{Information and Control}
{10(2)}
{189--208}
{1967}

\end{list}

\end{document}